\documentclass[11pt,a4paper]{article}
\usepackage[hyperref]{eacl2021}
\usepackage{times}
\usepackage{latexsym}

\usepackage{times}
\usepackage{url}
\usepackage{latexsym}
\usepackage{ifthen}

\usepackage[normalem]{ulem}
\useunder{\uline}{\ul}{}

\usepackage{amsmath}
\usepackage{amsfonts}
\usepackage{multirow}
\usepackage{booktabs}
\newcommand{\otoprule}{\midrule[\heavyrulewidth]}
\newcommand{\specialcell}[2][c]{%
  \begin{tabular}[#1]{@{}l@{}}#2\end{tabular}}

\usepackage[ruled,vlined]{algorithm2e}

\DeclareMathOperator*{\argmax}{arg\,max}

\aclfinalcopy 

\title{Implicit Unlikelihood Training: Improving Neural Text Generation with Reinforcement Learning}

\author{Evgeny Lagutin$^{2, 3, 4}$\space,\space Daniil Gavrilov$^{1,2,3}$\space,  Pavel Kalaidin$^{3, 5}$\thanks{\quad Work done while at VK.} \\
  $^1$VK, $^2$VK Lab, $^3$Moscow Institute of Physics and Technology, \\$^4$Skolkovo Innovation Center, Skolkovo Institute of Science and
Technology $^5$Tinkoff\\
  {lagutin.em@phystech.edu, daniil.gavrilov@vk.com, p.kalaydin@tinkoff.ai}
  }

\date{}

\begin{document}
\maketitle
\begin{abstract}
Likelihood training and maximization-based decoding result in dull and repetitive generated texts even when using powerful language models \cite{holtzman}. Adding a loss function for regularization was shown to improve text generation output by helping avoid unwanted properties, such as contradiction or repetition \cite{li}. In this work, we propose fine-tuning a language model by using policy gradient reinforcement learning, directly optimizing for better generation. We apply this approach to minimizing repetition in generated text, and show that, when combined with unlikelihood training \cite{welleck}, our method further reduces repetition without impacting the language model quality. We also evaluate other methods for improving generation at training and decoding time, and compare them using various metrics aimed at control for better text generation output.
\end{abstract}

\section{Introduction}
\label{intro}
Language models have become a subject of close attention in the Natural Language Processing field over the past few years. They are widely used not only for unsupervised pre-training, but also for text generation, such as what is implemented in dialogue systems \cite{roller}. While there are ongoing efforts to develop non-autoregressive models for language modeling, most current state-of-the-art approaches use the autoregressive method of generating text (i.e., word by word). \newcite{holtzman} showed that even powerful trained models with a high likelihood value for test data can output repetitive results. \newcite{schmidt} argues that the reason for that is train-test discrepancy and lack of generalization when running standard maximum likelihood estimation (MLE) training.

Unwanted repetition can be remedied at decoding and training time. Decoding methods focus on sampling techniques that generate less repetitive or incoherent samples, while other methods aim to improve model training to minimize the effects of degeneration. An effective method for reducing language model degeneration is unlikelihood training \cite{welleck}, where a regularization term forces the model to reduce the probability of generating a token that has already occurred in a sequence. \newcite{li} further explored this idea and showed that adding a loss function for regularization \textit{to avoid undesirable sequences} improves text generation not only by reducing repetition, but also by decreasing contradiction. \newcite{roller} reported that adding unlikelihood training also improves the humanness of generated text.

In this paper, we propose \textbf{Implicit Unlikelihood Training},
a method for regularizing output by fine-tuning a language model with policy gradient reinforcement learning to improve generation results. We apply this method for a repetition objective, and show that combining Implicit Unlikelihood Training with minimizing unlikelihood loss results in reduced repetition and perplexity. We also evaluate alternative approaches to improving generated texts in terms of repetitiveness, and compare these methods using a wide variety of metrics. 

The source code is available at: github.com/vklabmipt/implicit-unlikelihood-training.

\section{Related Work}

\subsection{Decoding Strategies}
\newcite{holtzman} observed that maximization-based decoding methods, such as top-k sampling \cite{fan}, beam search and its variations, can all lead to degeneration. They addressed this problem by using top-p (nucleus) sampling, proposing sampling from the top portion of the probability mass. \newcite{paulus} reported that ground-truth sentences for summarization tasks almost never contain the same trigram twice, and proposed the beam-blocking approach, where the decoder is forced to never output the same trigram more than once during testing. Penalized sampling \cite{keskar} works by discounting the scores of previously generated tokens. \newcite{martins} proposed preventing unlikely words from receiving any probability mass by using entmax sampling.

\subsection{Training Strategies}
\newcite{jiang} suggested that some tokens can be more difficult for a model to learn than others. These tokens are still under-learned after training, making their repetition more likely to happen. This issue is addressed by token loss dynamic reweighting (TLDR), which applies differentiable weights to individual token losses. Repetition can also be improved at training time by adding unlikelihood loss \cite{welleck,li} to regular likelihood loss. Unlikelihood training is aimed at decreasing the probability of previously generated tokens, and it was shown that it can outperform beam blocking and top-p sampling.

Coverage mechanisms \cite{tu,see} can also be used to reduce repetition. Adding pre-attention and highway connections was shown to decrease repetition for RNNs \cite{jiang}, while the architecture tweaks required for Transformers \cite{transformer} are still an open question.

\subsubsection{Unlikelihood Training}

Unlikelihood Training involves adding Unlikelihood Loss to lower the probability $p_{\theta}(c_i|x_{<t})$ of negative candidates $\mathcal{C}^{t}=\{c_1, c_2, ..., c_n\}$ at each timestamp:

\begin{equation}\label{eqn:ul_loss}
\mathcal{L}_{\mathrm{UL}}^{t}\left(p_{\theta}\left(\cdot | x_{<t}\right), \mathcal{C}^{t}\right)=-\sum_{c \in \mathcal{C}^{t}} \log \left(1-p_{\theta}\left(c | x_{<t}\right)\right)\text{.}
\end{equation}

We can construct the negative candidate set as $\mathcal{C}^{t} = \{x_{t-1}, x_{t_2}, ..., x_1\} \texttt{\textbackslash} \{x_t\}$ to improve generation results through reducing repetition. \newcite{welleck} also proposed using Sequence-Level Unlikelihood Loss on sampled continuations, where a sequence $(x_{t+1}, x_{t+2}, ..., x_{t + N})$ from a prefix $(x_{1}, x_{2}, ..., x_{t})$ is sampled first, and then the loss defined in Eq. \ref{eqn:ul_loss} for each $x_{t + i}$ (where $1 \leq i \leq N$) with negative examples $\mathcal{C}^{t + i}$ equal to $\{x_{t + i}\}$ if $x_{t + i}$ is a part of a repeating n-gram at a position before $t + i$ is minimized (see Algorithm \ref{algo:ul_step}
for details on Sequence-Level Unlikelihood Loss).
Fine-tuning a language model is then performed by
equally alternating between sequence-level unlikelihood and likelihood updates.

\subsection{Evaluation Metrics}
\label{metrics}

There are many metrics available for evaluating the performance (diversity or non-repetition) or quality of language models. These metrics include \textbf{perplexity}, the number of unique next-token predictions (\textbf{uniq}), \textbf{uniq-seq}, \textbf{rep/}$\mathbf{l}$, \textbf{wrep/}$\mathbf{l}$ \cite{welleck},  $\mathbf{\epsilon}$\textbf{-perplexity}, \textbf{sparsemax score} (sp), Jensen-Shannon divergence (\textbf{JSD}) \cite{martins}, and \textbf{DIMEN} \cite{jiang}. 

\textbf{Perplexity} is the metric used to evaluate language model quality. It is defined as $ppl(x) = p(x_1, x_2, ..., x_t)^{-\frac{1}{t}}$, where $x_1, x_2, ..., x_t$ is the sequence of tokens from test data. The lower the perplexity, the better the language model. 

\newcite{welleck} used a portion of duplicate n-grams in a generated sequence to measure \textbf{sequence repetition}: 
\begin{equation}
\label{eq:seq_rep}
    \textrm{seq\_rep\_n}(\mathbf{x}) = 1 - \frac{\textrm{\#uniq\_ngram}(\mathbf{x})}{\textrm{\#total\_ngram}(\mathbf{x})}.
\end{equation}
Higher repetition values mean that a language model tends to produce more repetitive output, which might appear less natural. Note that $0 \leq \textrm{seq\_rep\_n}(\mathbf{x}) < 1$. 

Same as \cite{welleck}, we controlled for the number of unique next-token predictions (\textbf{uniq}), as it was shown that generated texts are less diverse than those written by a human. We also used the number of unique tokens in continuations of validation or test prefixes (\textbf{uniq-seq}) as a measure of token distribution in generated text. \textbf{rep/}$\mathbf{l}$ is the fraction of next-token (top-1) predictions that occur in the previous $l$ tokens. \textbf{wrep/}$\mathbf{l}$ is a variant of rep/$l$ which only counts single token repetitions that are not equal to the ground truth next-token. We use 16,32,128,512 as $l$ and average the results to compute rep and wrep.

\newcite{martins} introduced $\mathbf{\epsilon}$\textbf{-perplexity} for computing perplexity of sparse distributions. The perplexity is smoothed by adding a small value of $\epsilon$ to all terms, followed by renormalization. They also introduced \textbf{sparsemax score} (sp) and \textbf{Jensen-Shannon divergence} (JSD) for evaluating quality and sparsity of probability distributions. For deterministic models, sparsemax score becomes word accuracy, and is bounded by 0 and 1. With JSD, the distance between the sparse or truncated distribution and the one-hot encoded ground truth distribution can be measured. It is used as a metric for language models using different decoding strategies. Unlike perplexity, JSD is bounded by $\log 2$.

\newcite{jiang} evaluate methods with a diversity metric based on n-grams (\textbf{DIMEN}). A high DIMEN score means that a set of generated sequences is diverse.

In this paper, we mainly focused on reducing sequence repetition (\textbf{seq\_rep\_n}) \cite{welleck}, which is a portion of duplicate n-grams in a generated sequence. Improving generation results by minimizing repetition should not significantly affect the \textbf{perplexity} of the language model.

\begin{table*}[h!]
    \centering
    \small
    \begin{tabular}{llllllll}
    \toprule                   
    \multirow{2}*{\textbf{method}} & \multirow{2}*{\textbf{ppl$\downarrow$}} & \multirow{2}*{\textbf{uniq$\uparrow$}}               &  \multicolumn{5}{c}{\textbf{seq\_rep\_4$\downarrow$}} \\ \cmidrule(lr){4-8}    
                  &             &               \textbf{top-k, k=1} &  \textbf{top-k, k=1}   &  \textbf{top-k, k=3}   & \textbf{top-k, k=8}   &  \textbf{top-p, p=0.3}   &  \textbf{top-p, p=0.9}   \\

    \otoprule
    \multicolumn{8}{c}{small GPT-2 model, 0.5 update rate} \\
    \otoprule
     PG, c=3           & 19.409$\pm$.195 & 11259$\pm$52  & .032$\pm$.014     & .024$\pm$.010      & .017$\pm$.006     & .029$\pm$.012       & .007$\pm$.001       \\
     \midrule
     PG + UT, c=3           & 19.344$\pm$.034 & 11279$\pm$53  & .010$\pm$.002      & .008$\pm$.001     & .008$\pm$.000       & .012$\pm$.002       & .007$\pm$.001       \\
     \midrule
     i-UT, c=3   & 19.182$\pm$.032 & 11308$\pm$73  & .009$\pm$.002     & .008$\pm$.001     & .008$\pm$.001     & .012$\pm$.002       & .007$\pm$.001       \\
     i-UT, c=9   & 19.302$\pm$.051          & 11297$\pm$36     & .006$\pm$.001 & .006$\pm$.001 & .007$\pm$.002 & .009$\pm$.002 & .006$\pm$.001 \\
     i-UT, c=15   & \textbf{19.170$\pm$.123}  & \textbf{11432$\pm$34} & .007$\pm$.005     & .007$\pm$.002     & \textbf{.007$\pm$.001}     & .010$\pm$.003        & .007$\pm$.001       \\
     i-UT, c=30  & 19.504$\pm$.065 & 11427$\pm$66  & \textbf{.005$\pm$.000}       & \textbf{.005$\pm$.001}     & \textbf{.007$\pm$.001}     & \textbf{.005$\pm$.001}       & \textbf{.006$\pm$.001}       \\
     \midrule
     UT             & 19.442$\pm$.085 & 11210$\pm$47  & .056$\pm$.033     & .011$\pm$.002     & .008$\pm$.001     & .014$\pm$.003       & \textbf{.006$\pm$.001}       \\
     \midrule
     UT w/ warmup & 19.347$\pm$.065 & 11295$\pm$4  & .055$\pm$.016     & .010$\pm$.001      & .008$\pm$.001     & .014$\pm$.002       & \textbf{.006$\pm$.001}       \\
     \otoprule%
     
     \multicolumn{8}{c}{small GPT-2 model, 0.25 update rate} \\

    \otoprule%
     PG, c=3     & 18.516$\pm$.086 & 11492$\pm$32   & .060$\pm$.011      & .035$\pm$.005     & .022$\pm$.003     & .043$\pm$.006       & .009$\pm$.001       \\
     PG, c=9     & 18.638$\pm$.022 & 11496$\pm$21 & .025$\pm$.011     & .022$\pm$.008     & .018$\pm$.005     & .025$\pm$.008       & .009$\pm$.001       \\
     PG, c=15     & 18.696$\pm$.062 & 11533$\pm$3   & .022$\pm$.002     & .019$\pm$.001     & .016$\pm$.000       & .021$\pm$.001       & .008$\pm$.001       \\
     PG, c=30    & 19.026$\pm$.099 & 11487$\pm$36 & .031$\pm$.021     & .027$\pm$.017     & .020$\pm$.010       & .027$\pm$.016       & .008$\pm$.001       \\
     \midrule
     i-UT, c=3  & 18.504$\pm$.059 & 11519$\pm$21   & .020$\pm$.002      & .014$\pm$.001     & .012$\pm$.001     & .017$\pm$.002       & \textbf{.007$\pm$.001}       \\
     
     i-UT, c=9  & \textbf{18.487$\pm$.022} & 11552$\pm$51  & .011$\pm$.003     & .011$\pm$.003     & .011$\pm$.002     & .016$\pm$.003       & \textbf{.007$\pm$.000}         \\
     i-UT, c=15  & 18.590$\pm$.065  & 11504$\pm$24   & \textbf{.008$\pm$.001}     & \textbf{.008$\pm$.001}     & \textbf{.008$\pm$.001}     & .011$\pm$.002       & \textbf{.007$\pm$.001}       \\
     i-UT, c=30 & 18.733$\pm$.082 & \textbf{11558$\pm$27}   & \textbf{.007$\pm$.001}     & \textbf{.008$\pm$.001}     & .009$\pm$.001     & \textbf{.010$\pm$.001}        & .008$\pm$.000         \\ 
     \midrule
     UT            & 18.764$\pm$.164 & 11377$\pm$55   & .055$\pm$.013     & .011$\pm$.002     & .009$\pm$.001     & .016$\pm$.003       & \textbf{.007$\pm$.001}       \\
    \otoprule%
    
    \multicolumn{8}{c}{medium GPT-2 model, 0.5 update rate} \\
    
        \otoprule%
     i-UT, c=3 & \textbf{13.620$\pm$.015} & 12437$\pm$29 & .013$\pm$.002     & .010$\pm$.002      & .010$\pm$.001      & .011$\pm$.002       & .008$\pm$.001      \\ 
     
      i-UT, c=15  & 13.669$\pm$.018 & 12355$\pm$3    & \textbf{.011$\pm$.003}       & .010$\pm$.002        & .011$\pm$.002       & .012$\pm$.001         & .009$\pm$.000          \\
      
    i-UT, c=30 & 13.785$\pm$.004   & \textbf{13319$\pm$898} & \textbf{.011$\pm$.003}       & .009$\pm$.003       & .011$\pm$.001       & \textbf{.010$\pm$.004}          & \textbf{.008$\pm$.000} \\

     \midrule
     UT      & 13.710$\pm$.009 & 12386$\pm$29 & .024$\pm$.003     & \textbf{.008$\pm$.001}     & \textbf{.009$\pm$.001}     & .013$\pm$.001       & .008$\pm$.001       \\
    \bottomrule
    \end{tabular}
    \caption{Repetition on small and medium GPT-2 models. Validation data was used as sampling prefixes for evaluating the metrics.}
    \label{tab:small_seq_rep_0.5_valid}
\end{table*}

\section{Implicit Unlikelihood Training}

\newcite{li} showed that Unlikelihood Training can be employed as a general framework for reducing the likelihood of undesirable text generation results through training on negative examples. However, we argue that, in some cases, it could be difficult to construct negative samples for specific types of Unlikelihood Loss\footnote{This can include reducing the toxicity or bias level of generated sequences by using a score from an external classifier.}. To address this issue, we propose extending Unlikelihood Training with policy gradient reinforcement learning, which does not need explicitly created negative samples.

We chose to test this approach for repetition as the most widely considered property of neural text degeneration. To directly minimize repetition (see Equation \ref{eq:seq_rep}) for sequence $x$, we define the reward as $R = 1 - \textrm{seq\_rep\_n}(\mathbf{x})$ with $n=4$. We alternated between maximizing the reward $R$, optimizing the likelihood of training data, and Sequence-Level Unlikelihood Loss (see \ref{algo:alternating}, \ref{algo:pg_step} and \ref{algo:ul_step} for details on the process of Implicit Unlikelihood Training and policy gradient update).

\begin{algorithm}[h!]
\small
\SetAlgoLined
\KwIn{update rate $r$, total number of updates $N$}
 \For{i = 1 to N}{
    sample $x \sim \text{U}[0,1]$ \\
    \eIf{$x < r$}{
        sample $y \sim \text{U}[0,1]$ \\
        \eIf{$y < 0.5$}{
            do a policy gradient update
            }{
            do a sequence-level unlikelihood update
            }
        }{
        do a MLE update
        }
  }
\caption{i-UT: alternating between MLE, UT and PG updates}
\label{algo:alternating}
\end{algorithm}

\begin{algorithm}[h!]
\small
\SetAlgoLined
\KwIn{LM $\theta$, $m$ prefixes $\mathcal{D}_{m} = \{(x_1^{(j)}, \dots, x_k^{(j)}),~j=\overline{1..m}\}$, continuation length $T$}
\KwOut{loss $\mathcal{L}(\theta, \mathcal{D}_m)$}
\For{$j$ = $1$ to $m$} {
 \For{$t$ = $k+1$ to  $k+T$}{
    Get  $p_{\theta}(\cdot \mid x_{<t}^{(j)})$
    
    $x_{t}^{(j)} = \underset{x \in \mathcal{V}}{\argmax}~{p_{\theta}(\cdot \mid x_{<t}^{(j)})}$ 
 }
}
 \For{$j$ = 1 to $m$}{
    $R_j = 1 - \textrm{seq\_rep\_n}(\mathbf{x}^{(j)})$ 
 }
 $b(\mathbf{x}^{(1)},\dots, \mathbf{x}^{(m)}) = \sum\limits_{j=1}^{m} R_j$ 
 
 \For{$j$ = 1 to $m$}{
    $\Psi_j = R_j - b(\mathbf{x}^{(1)},\dots, \mathbf{x}^{(m)})\cdot\frac{1}{m}$ 
 }
 $\mathcal {L}(\theta, \mathcal{D}_m) = -\dfrac{1}{m}\sum\limits_{j=1}^{m}\Psi_{j} \cdot \dfrac{1}{T}\sum\limits_{t=k+1}^{k+T}\log p_\theta(x_{t}^{(j)}~|~x_{<{t}}^{(j)}) $
 \caption{Policy Gradient Update}
\label{algo:pg_step}
\end{algorithm}

\begin{algorithm}[h!]
\small
\SetAlgoLined
\KwIn{LM $\theta$, batch of prefixes $\mathcal{D}_{m} = \{(x_1^{(j)}, \dots, x_k^{(j)}),~j=\overline{1..m}\}$, continuation length $T$}
\KwOut{loss $\mathcal{L}(\theta, \mathcal{D}_{m})$}
\For{$j$ = $1$ to $m$} {
 \For{$t$ = $k+1$ to  $k+T$}{
    Get  $p_{\theta}(\cdot \mid x_{<t}^{(j)})$
    
    $x_{t}^{(j)} = \underset{x \in \mathcal{V}}{\argmax}~{p_{\theta}(\cdot \mid x_{<t}^{(j)})}$ 
 }
}
\For {$j$ = 1 to $m$}{
  \For {$t$ = $k+1$ to $k+T$}{
    \eIf{$\left(x_{t-i}, \ldots, x_{t}, \ldots, x_{t+h}\right) \in x^{(j)}_{<t-i} \text { for any }(h-i)=n, i \leq n \leq h$}{
         $\mathcal{C}_{\text{repeat-n}}^{t}(\mathbf{x}^{(j)}) =\left\{x_{t}\right\}$
    }{
         $\mathcal{C}_{\text{repeat-n}}^{t}(\mathbf{x}^{(j)}) =\emptyset$
    }
  }
}
$\mathcal{L}(\theta, \mathcal{D}_{m}) = \dfrac{1}{m}\sum\limits_{j=1}^{m}\dfrac{1}{T}\sum\limits_{t=k+1}^{k+T}\mathcal{L}^{t}_{UL}\left(p_{\theta}(\cdot \mid x^{(j)}_{<t}), \mathcal{C}^t_{\text{repeat-n}}\right)$ 
     
 \caption{Sequence-Level Unlikelihood Update}
 \label{algo:ul_step}
\end{algorithm}

\section{Experiment Details}
\subsection{Setup}

We fine-tuned small and medium GPT-2 models (175M and 345M parameters, respectively) \cite{radford} on the WikiText-103 dataset \cite{merity}.

Our experiments consisted of alternating between three types of updates: Maximizing Likelihood (MLE), minimizing Sequence-Level Unlikelihood Loss (UL), and minimizing repetition with policy gradient reinforcement learning (PG). The first approach is a plain \textbf{MLE} update, for which we do not use any specific methods for reducing repetition in samples. We also experimented with Unlikelihood Training (\textbf{UT}), which involves alternating between MLE and UL updates (see Algorithm \ref{algo:alternating} for details).

\begin{table*}[h!]
    \small
    \centering
        \begin{tabular}{llrrrrrr}
    \toprule
                         \textbf{sampling} &  \textbf{method} &    \textbf{uniq$\uparrow$} &   \textbf{rep$\downarrow$} &   \textbf{wrep$\downarrow$} &   \textbf{$\epsilon-$ppl/ppl$\downarrow$} &   \textbf{JSD$\downarrow$} &
                         \textbf{sp$\uparrow$} \\
    \otoprule
     \multirow{3}*{\specialcell[t]{softmax \\ $\epsilon=0$}} 
     & MLE   & 19932   & .373 &  \textbf{.174} &  13.830  &    .382 & .680 \\
     & i-UT, c=3  & \textbf{20493}  & .372 &  \textbf{.174} &  \underline{\textbf{13.161}} &    \underline{.379}  & \underline{\textbf{.683}} \\
     & UT      & 20419 & .371 &  \textbf{.174} &  13.266 &    .380 & .682  \\
    \midrule
     \multirow{3}*{\specialcell[t]{greedy \\ $\epsilon=2\times10^{-5}$}} 
     & MLE    & 12639   & .489 &  .230  & 539.930  &    .358 & .483 \\
     & i-UT, c=3   & 12859 & .488 &  \underline{.228} & \underline{511.478} &    \underline{\textbf{.355}} & \underline{.488}\\
     & UT       & 12826 & .488 &  \underline{.228} & 517.316 &    .356 & .487\\
     \midrule
      \multirow{3}*{\specialcell[t]{top-k, k=10 \\ $\epsilon=8\times10^{-6}$}} 
      & MLE     & 14326   & .436 &  .220  &  50.003 &    .358 & .668\\
     & i-UT, c=3      & 14731 & .435 &  \underline{.219} &  \underline{46.885} &    \underline{\textbf{.355}} & \underline{.672} \\
     & UT        & 14648 & .435 &  \underline{.219} &  47.248 &    .356 & .671 \\
    \midrule
     \multirow{3}*{\specialcell[t]{top-p, p=0.9 \\ $\epsilon=2\times10^{-6}$}} 
     & MLE   & 17589   & .395 &  .186 &  19.689 &    .371 & .678\\
     & i-UT, c=3   & 18116   & .394 &  .185 &  \underline{18.662} &    \underline{.368} & \underline{.682} \\
     & UT       & 17964 &.393 &  .186 &  18.782 &    .369 & .681 \\
     \midrule
     \specialcell[t]{$\alpha$-entmax ($\alpha=1.2$) \\ $\epsilon = 1\times10^{-6}$} & \specialcell[t]{$\alpha$-entmax loss \\($\alpha=1.2$)} &  19942 &  \textbf{.370} &  .176 &  15.124 &    .389 & .680 \\
    \bottomrule
    \end{tabular}
        \caption{Repetition on the medium GPT-2 model with a 0.5 update rate. Regular perplexity is reported for softmax sampling. Test data is used as sampling prefixes for evaluating the metrics.}
    \label{tab:sampling}
\end{table*}

In policy gradient experiments, we trained models in three different scenarios: a plain \textbf{PG} for which we alternated MLE and PG updates; a combined \textbf{PG + UT} approach, where we alternated between maximizing the likelihood and minimizing the sum of policy gradient and unlikelihood losses; and finally, the proposed Implicit Unlikelihood Training (\textbf{i-UT}), which consisted of alternating between MLE, UL and PG updates (see Algorithms \ref{algo:alternating}, \ref{algo:ul_step}, and \ref{algo:pg_step}). We used 0.25 and 0.5 alternating update rates for the small GPT-2 model, and $r$ equal to $0.5$ for the medium GPT-2 model. 

Full optimization details are provided in \textbf{Appendix \ref{appendix:opt}. Optimization Details}.

\subsection{Evaluation}

We used top-k and top-p samplings with different $k$ and $p$ to evaluate sequence repetition\footnote{Note that top-k with $k=1$ is greedy sampling by definition.} (seq\_rep\_4) for the described approaches. For these experiments, we used validation data to evaluate perplexity and to generate the sampling prefixes for evaluating uniq and seq\_rep\_4 metrics. The number of unique tokens (uniq) was evaluated using greedy sampling. We also evaluated the proposed method with rep, wrep, and JSD metrics using different sampling methods on test data, and compared it with other related approaches (MLE, UT, entmax). We repeated each experiment 5 times and reported the mean and standard deviation values of the measurements. 

More experiments and their results are described in \textbf{Appendix \ref{appendix:experriments}. Experiments}.

\section{Results}
We showed that Implicit Unlikelihood Training is a competitive approach that outperforms other methods in sequence repetition when fine-tuning small and medium GPT-2 models (see Table  \ref{tab:small_seq_rep_0.5_valid}) on most variants of top-k and top-p sampling, while maintaining the lowest perplexity and the highest count of unique tokens generated. This approach also achieved better results than training with entmax loss and other related approaches, using a different range of sampling methods (see Table \ref{tab:sampling}), with the only exception being the rep metric, where entmax performed similar to i-UT.

Samples of generated outputs are provided in Tables \ref{app:samples}, \ref{app:samples2} in Appendix. 

\subsection{Negative Results}

Our results showed that all sampling methods, other than greedy sampling, led to worse convergence of the seq\_rep\_4 metric.


We experimented with using the Proximal Policy Optimization algorithm \cite{ppo} for PG update, but faced unstable validation perplexity behavior during training, and did not obtain any comparable results.

Another unsuccessful direction of our experiments was substituting the estimation of the reward calculated on the full sequence with the reward put on each token separately. We tried using two variants of the binary reward function: does the current n-gram appear first time in the text, and does the current n-gram appear in the following part of the text. We experimented with advantage estimation by using a value function estimator, and without it by using pure rewards. In the former case, we adjusted different values of $\lambda$ and $\gamma$ for the Generalized Advantage Estimation algorithm \cite{gae}, and in the latter, we used a general discounted future reward. We observed that the approach of estimating a single reward for a whole sequence and subtracting a baseline value to reduce the variance of the gradient estimation performed best.

\section{Future Work}
The described and evaluated reinforcement learning framework makes it possible to optimize text generation for any objective. In future work, we intend to test the approach not only for repetition, but also for various other metrics, such as the toxicity level or bias of generated text.

\section{Acknowledgements}
The authors are grateful to anonymous reviewers for valuable comments and to Viktoriia Loginova and David Prince for proofreading.

\bibliography{anthology,eacl2021}
\bibliographystyle{acl_natbib}

\appendix
\clearpage

\section{Appendices}

\subsection{Optimization Details}
\label{appendix:opt}
For likelihood update, we evaluated the likelihood of a sequence of tokens with lengths equal to $300$. For both UL and PG updates, we formed prefixes using a sequence of $300$ tokens to form $6$ sequences with lengths equal to $50$. We then used these prefixes to sample sequences with a maximum length of $100$ tokens.

For optimization, we used the Adam optimizer \cite{kingma2014adam} with a learning rate of $6.25 \times 10^{-5}$. Similar to \cite{welleck} and \cite{martins}, we did no warmup steps for UT and $\alpha-$entmax training. For i-UT, we did 500 linear warm-up steps . After warm-up steps, we linearly decayed the learning rate to zero. 

In all our experiments, we fine-tuned language models for 5000 total updates. 

Once training was complete, we selected a checkpoint with the least validation perplexity obtained during training. This is the last checkpoint in most of our experiments, which means that general log-likelihood loss converges.

As shown in Algorithm \ref{algo:alternating}, we equally alternated between UL and PG updates. We also found that reducing unlikelihood update rate to 0.25 may also be effective, taking twice less time (see Table \ref{tab:small_seq_rep_0.5_valid}). The parameters $\epsilon$ for $\epsilon-\text{ppl}$ and $\alpha$ for $\alpha-\text{entmax}$ training were taken from \cite{martins} (except $\epsilon$ for $\alpha-\text{entmax}$, which we set to $1\times10^{-6}$). 

We conducted coefficient search on our policy gradient loss with $c = \{3, 9, 15, 30\}$ for the small GPT-2 model, and $c = \{3, 15, 30\}$ for medium GPT-2 model. We chose the best models based on the results on the validation set, and also reported the metrics on the test set.

\subsection{Experiments}
\label{appendix:experriments}
We evaluated DIMEN and uniq-seq for UT and i-UT methods, applied to small and medium GPT-2 models using different sampling methods for DIMEN, and greedy sampling for evaluation of uniq-seq. In this experiment, we observed that Implicit Unlikelihood Training performed better or equal to Unlikelihood Training with different sampling methods measured by the DIMEN metric, having a significantly better value of uniq-seq (see Table 7).

We also evaluated sequence repetition with beam-search sampling for MLE, UT, and i-UT methods for both small and medium GPT-2 models, using validation data to form sampling prefixes. When sampling with beam search, we found that Implicit Unlikelihood Training produced better results than Unlikelihood Training (see Table \ref{tab:beam_search}). 

For greedy sampling with small GPT-2 model, we evaluated sequence repetition, wrep, uniq, and perplexity. We used test data to evaluate the perplexity, and to form sampling prefixes for other methods. We observed that MLE, UT, and i-UT methods had similar performance in terms of repetition using greedy sampling, while i-UT still had the best number of unique tokens (see Table \ref{tab:small_seq_rep_0.5_test}).

Finally, we evaluated the TLDR method using both sequence repetition and DIMEN metrics (see Tables \ref{tab:tldr_valid}, \ref{tab:tldr_valid_2}). In our experimental setup, TLDR performed on par with MLE approach.

\begin{table*}[h!]
\parbox{.37\linewidth}{
   \small
    \centering
    \begin{tabular}{lcc|c}
    \toprule
           & \multicolumn{3}{c}{\textbf{seq\_rep\_4$\downarrow$}}   \\
           \cmidrule(lr){2-4}
        GPT-2:   & \multicolumn{2}{c|}{small} & medium \\
            \cmidrule(lr){1-4}
            
        seq rate:   & .5 & .25 & .5 \\
    \otoprule
     MLE   & .67$\pm$.01 & .67$\pm$.01 & .64$\pm$.01         \\
     i-UT      & \textbf{.03$\pm$.04} & \textbf{.08$\pm$.03} & \textbf{.08$\pm$.02}      \\
     UT  & .08$\pm$.03 & .14$\pm$.02 & .13$\pm$.02       \\
    \bottomrule
    \end{tabular}
    \caption{Repetition with Beam Search. Validation data is used as sampling prefixes for evaluating the metrics. We reported the results of i-UT model with the value of $c$ for which we had the best validation perplexity.}
    \label{tab:beam_search}
    }
\hfill
\parbox{.55\linewidth}{
    \centering
    \small
    \begin{tabular}{llllll}
    \toprule
          \textbf{method}  & \textbf{ppl$\downarrow$}          & \textbf{rep$\downarrow$}    & \textbf{wrep$\downarrow$}   & \textbf{uniq$\uparrow$}            \\
    \otoprule
      MLE   & 17.94$\pm$.03     & .504$\pm$.001 & .252$\pm$.001 & 11790$\pm$92 \\
 i-UT, c=15 & 18.54$\pm$.13 &   .504$\pm$.001 &.254$\pm$.0   & 11847$\pm$32  \\
 UT    & 18.76$\pm$.09 &   .503$\pm$.001 & .253$\pm$.001 & 11597$\pm$52  \\
    \bottomrule
    \end{tabular}
    \caption{Repetition on small GPT-2 model, UT and i-UT 0.5 update rate with greedy sampling. Test data is used as sampling prefixes for evaluating the metrics.}
    \label{tab:small_seq_rep_0.5_test}
    }
\end{table*}

\begin{table*}[h!]
\centering
    \small
    \begin{tabular}{lllllllll}
    \toprule
         \multirow{2}*{\textbf{method}} & \multirow{2}*{\textbf{ppl$\downarrow$}} & \multirow{2}*{\textbf{uniq$\uparrow$}}               & 
          \multicolumn{4}{c}{\textbf{seq\_rep\_4$\downarrow$}} \\ \cmidrule(lr){4-8}    
          &   &   \textbf{top-k, k=1}   &  \textbf{top-k, k=1}   &  \textbf{top-k, k=3}   & \textbf{top-k, k=8}   &  \textbf{top-p, p=0.3}   &  \textbf{top-p, p=0.9}   \\
    \otoprule
      MLE   & 18.611$\pm$.088 & 11361$\pm$37 & .550$\pm$.035        & .131$\pm$.004       & .054$\pm$.001       & .233$\pm$.005         & .013$\pm$.001         \\
 TLDR  & 18.713$\pm$.059 & 11322$\pm$42  & .514$\pm$.012       & .118$\pm$.004       & .047$\pm$.002       & .241$\pm$.005         & .011$\pm$.001         \\
    \bottomrule
    \end{tabular}
    \caption{Repetition on small GPT-2 model for TLDR and MLE approaches. Validation data is used as sampling prefixes for evaluating the metrics.}
    \label{tab:tldr_valid}
\end{table*}

\begin{table*}[h!]
    \small
    \centering
    \begin{tabular}{lllllll}
    \toprule
         \multirow{2}*{\textbf{method}} &  \multirow{2}*{\textbf{uniq-seq$\uparrow$}}               & 
          
          \multicolumn{5}{c}{\textbf{DIMEN$\uparrow$}} \\ \cmidrule(lr){3-7}    
                      &  \textbf{top-k, k=1} &    \textbf{top-k, k=1}   &  \textbf{top-k, k=3}   & \textbf{top-k, k=8}   &  \textbf{top-p, p=0.3}   &  \textbf{top-p, p=0.9}   \\
    \otoprule
MLE  & 8074.75$\pm$182.398 & .365$\pm$.023 & .685$\pm$.003 & .778$\pm$.001 & .599$\pm$.004 & .869$\pm$.001 \\
 TLDR & 8122.0$\pm$112.018  & .390$\pm$.008  & .692$\pm$.004 & .781$\pm$.002 & .590$\pm$.004  & .867$\pm$.001 \\
     \bottomrule
    \end{tabular}
    \caption{Diversity on small GPT-2 model for TLDR and MLE approaches. Validation data is used as sampling prefixes for evaluating the metrics.}
    \label{tab:tldr_valid_2}
\end{table*}

\fi

\begin{table*}[h!]
   \small
   \centering
    \begin{tabular}{lllllll}
    \toprule
    
  \multirow{2}*{\textbf{method}} & \multirow{2}*{\textbf{uniq-seq$\uparrow$}}               &  \multicolumn{5}{c}{\textbf{DIMEN$\uparrow$}} \\ \cmidrule(lr){3-7}    
                  &  \textbf{top-k, k=1} &  \textbf{top-k, k=1}   &  \textbf{top-k, k=3}   & \textbf{top-k, k=8}   &  \textbf{top-p, p=0.3}   &  \textbf{top-p, p=0.9}   \\
                  
    \otoprule
    \multicolumn{7}{c}{small GPT-2 model, 0.5 update rate} \\
    \otoprule
     i-UT, c=3  & \textbf{11340$\pm$703}  & .855$\pm$.008 & .859$\pm$.005 & .859$\pm$.003 & .834$\pm$.005 & .880$\pm$.003  \\
     i-UT, c=9  & 10547$\pm$533	          & \textbf{.881$\pm$.014} & .876$\pm$.003 & .870$\pm$.004  & .852$\pm$.0   & \textbf{.881$\pm$.003} \\
     i-UT, c=15  & 10621$\pm$489 & .863$\pm$.023 & .867$\pm$.013 & .864$\pm$.006 & .855$\pm$.014 & \textbf{.881$\pm$.002} \\
     i-UT, c=30 & 10771$\pm$265  & \textbf{.881$\pm$.009} & \textbf{.880$\pm$.008}  & \textbf{.871$\pm$.007} & \textbf{.880$\pm$.012}  & .880$\pm$.003  \\
     \midrule
     UT          & 9651$\pm$436   & .785$\pm$.044 & .847$\pm$.012 & .856$\pm$.005 & .831$\pm$.011 & .880$\pm$.002  \\
    \otoprule
    \multicolumn{7}{c}{small GPT-2 model, 0.25 update rate} \\
    \otoprule
     i-UT, c=3  & 10885$\pm$375   & .817$\pm$.006 & .834$\pm$.003 & .845$\pm$.002 & .819$\pm$.003 & .878$\pm$.002 \\
     i-UT, c=9  & 11208$\pm$884 & .852$\pm$.014 & .849$\pm$.01  & .851$\pm$.007 & .828$\pm$.01  & .879$\pm$.001 \\
     i-UT, c=15  & \textbf{11966$\pm$1019}  & .848$\pm$.012 & .854$\pm$.006 & .856$\pm$.004 & .837$\pm$.006 & .879$\pm$.001 \\
     i-UT, c=30 & 10418$\pm$762   & \textbf{.874$\pm$.012} & \textbf{.871$\pm$.01}  & \textbf{.863$\pm$.007} & \textbf{.863$\pm$.017} & .878$\pm$.002 \\
     \midrule
     UT          & 9696$\pm$441    & .778$\pm$.017 & .843$\pm$.007 & .854$\pm$.003 & .827$\pm$.008 & \textbf{.880$\pm$.002}  \\
    \otoprule
        \multicolumn{7}{c}{medium GPT-2 model, 0.5 update rate} \\
    \otoprule
     i-UT, c=3   & \textbf{12900$\pm$303} & .843$\pm$.006 & .854$\pm$.005 & .857$\pm$.004 & .845$\pm$.006 & \textbf{.877$\pm$.002} \\
     i-UT, c=15  & 13849$\pm$0     & .869$\pm$.003 & .861$\pm$.002 & .857$\pm$.003 & .854$\pm$.0   & .875$\pm$.001 \\
     i-UT, c=30 & 11748$\pm$340   & \textbf{.880$\pm$.004}  & \textbf{.871$\pm$.01}  & \textbf{.861$\pm$.006} & \textbf{.867$\pm$.009} & \textbf{.877$\pm$.002} \\
     \midrule
     UT      & 12649$\pm$262 & .837$\pm$.006 & .869$\pm$.004 & .860$\pm$.003  & .842$\pm$.005 & .875$\pm$.001 \\
     \bottomrule
    \end{tabular}
    \label{tab:dimen-valid}
    \caption{Diversity on small and medium GPT-2 models. Validation data is used as sampling prefixes for evaluating the metrics. }
\end{table*}

\clearpage






\label{app:negative}
\clearpage


\begin{table*}[h]
\small
\center
\begin{tabular}{llr}
    \otoprule

\multicolumn{2}{l}{} & seq\_rep\_4 \\
    \midrule

prefix: & \begin{tabular}[c]{@{}l@{}}Air Vice Marshal Frank Headlam, CB, CBE ( 15 July 1914 – 23 December
1976 ) \\ was a senior  commander in the Royal Australian Air Force ( RAAF). \\Born and educated in Tasmania, he joined the RAAF in\end{tabular} &  \\

   \otoprule
    \multicolumn{3}{c}{medium GPT-2 model, 0.5 update rate, top-k, k=1} \\
    \otoprule

MLE & \begin{tabular}[c]{@{}l@{}}1914, and was promoted to the rank of CB. \\ He was promoted to the rank
of CBE in 1916, and was promoted to the rank of CB. \\ He was promoted to
the rank of CBE in 1917, and was promoted to the rank of CB. \\ He was
promoted to the rank of CBE in 1918, and was promoted to the rank of CB.
\\ He was promoted to the rank of CBE in 1919, and was promoted to the rank
of CB. \\He was promoted to\end{tabular} & .6535 \\\\

UT & \begin{tabular}[c]{@{}l@{}}1914 and was promoted to the rank of lieutenant in 1916. He was
promoted \\ to commandant in the Australian Air Force ( R @-@ Special
Operations Command ( C-MEDF ) \\ in spare parts for related acronym changed
to $<$unk$>$ ) Special Operations Command \\( Special Commander in Command by
extension ) being a technical specialist \\ in communications equipment and
equipment related to intelligence gathering \\ and analysis, and related
related related jargon, and related related related equipment \\ related to
logistics and logistics management.
 Headlam\end{tabular} & .0288 \\\\

i-UT & \begin{tabular}[c]{@{}l@{}}1914 and served in the Australian Army Air Corps ( AAF ) from 1915 to
1918. \\ Headlam was promoted to the rank of lieutenant colonel in 1918 and
promoted \\ to the rank and title of commander in the Pacific theatre in
the Pacific theatre in \\ ( R \& R ) $<$unk$>$ in order to be eligible for the
Medal of Honor for gallantry in \\ action in the Battle of the Gulf, but
the award was not awarded until the end of \\ the war. Headley was awarded
the Distinguished Service\end{tabular} & .0189 \\
    
    \otoprule
    \multicolumn{3}{c}{medium GPT-2 model, 0.5 update rate, top-k, k=8} \\
    \otoprule
    
    MLE & \begin{tabular}[c]{@{}l@{}}1917 at the age of twenty @-@ seven, and served at both the Western and Eastern \\Airports, and the Western Reserve Air Training Center. He was awarded a\\ Distinguished Service Order in 1917, and a Military Medal for his service \\in World War I, and a Distinguished Service Order for his services during the \\Great Depression. In 1920 he was appointed as Chief of the Staff of the RAAF \\and served as the RAAF's Chief of Staff. In 1921 he was appointed a\end{tabular} & .0211 \\\\

UT & \begin{tabular}[c]{@{}l@{}}1939 as a private with the rank of Captain, serving on board Australian carriers, such as \\the RAN @-@ class battleships HMAS Melbourne and HMAS Sydney. In the early years\\ of the war the Australian navy's air force, with the exception of the RAN, \\was under the command and control of Rear @-@ Admiral Robert W. Campbell, the commander \\of the RAAF in Australia. During the Second World War, Headlam became a member\\ of the RAN\end{tabular} & .0105 \\\\

i-UT & \begin{tabular}[c]{@{}l@{}}1917 as a private and served in the Second World War. Headlam became the second man \\in the RAAF to be awarded the Victoria Cross in December 1941 for valour and gallantry \\in a landmine @-@ exploding raid on a German convoy in the Gulf of Mexico.\\ 
  Headlam became Commander of the Royal Australian Air Force in July 1943 after serving as\\ a Lieutenant in the RAAF from 1917 to 1919 and then as Commander of the RAAF in the\\ First World War and\end{tabular} & .0104 \\

\bottomrule
\end{tabular}
    \caption{Generation Samples} \label{app:samples}
\end{table*}

\begin{table*}[h]
\small
\center
\begin{tabular}{llr}
    \otoprule

\multicolumn{2}{l}{} & seq\_rep\_4 \\
    \midrule

prefix: & \begin{tabular}[c]{@{}l@{}}Air Vice Marshal Frank Headlam, CB, CBE ( 15 July 1914 – 23 December
1976 ) \\ was a senior  commander in the Royal Australian Air Force ( RAAF). \\Born and educated in Tasmania, he joined the RAAF in\end{tabular} &  \\

   \otoprule
    \multicolumn{3}{c}{medium GPT-2 model, 0.5 update rate, top-p, k=0.9} \\
    \otoprule

MLE & \begin{tabular}[c]{@{}l@{}}1940 as a gunnery instructor and officer at Hobson Air Station, Melbourne. \\On 3 January 1940 he led a small squad that formed part of the Caraboo Air Force \\( the first to be taught at a Melbourne naval base ) that did air training \\and drills at the Haro Air Force Base. Headlam was later commanding the \\200th ( First United States Army Air Service Group, Unit A ) and Unit B ( the second \\to be taught at a Melbourne naval base ). In the\end{tabular} & .0526 \\\\

UT & \begin{tabular}[c]{@{}l@{}}December 1914 at the age of 15 years and 177 days. Headlam served with the RAAF and the \\Italian Free Fire Division before leaving in the middle of the War in October 1917 \\to become a field marshal of the Melbourne Transport Service. In 1919 he was nominated \\for a Distinguished Service Order ( DSO ) and he retired as a Field Marshal in December \\1918. After the War he served in the RAAF as a field marshal until leaving to become \\an air force field marshal in
 Headlam\end{tabular} & .0104 \\\\

i-UT & \begin{tabular}[c]{@{}l@{}}October 1914 as a radio operator and served with distinction until his discharge from the \\RAAF in December. 
  In February 1915, Headlam was appointed to the role of \\President of the RAC, and a policy advisory officer to the PM. He was promoted \\to the rank of Chief of the Air Staff, and underlined that he was a major and not \\to have the responsibility of determining the combatant classes ( like the \\Royal Australian Flying Corps ). When the RAC was created in\end{tabular} & .0 \\

\bottomrule

\end{tabular}

\caption{Generation Samples} \label{app:samples2}
\end{table*}

\end{document}